%% file: paper.tex
\documentclass[runningheads]{llncs}
\usepackage[T1]{fontenc}
\usepackage{graphicx}
\usepackage{amsmath}
\usepackage{xcolor}
\usepackage{color}
\usepackage{subcaption}
\usepackage{tabularx}
\usepackage{multirow}
\usepackage{booktabs}

\usepackage{siunitx}
\usepackage{graphicx}
\usepackage[pagebackref=true,breaklinks=true]{hyperref}
\hypersetup{
     colorlinks,
     linkcolor={red!75!black},
     citecolor={blue!75!black},
     urlcolor={blue!75!black},
}

\usepackage[capitalise]{cleveref}

\usepackage{microtype}

\usepackage{xspace}
\makeatletter
\DeclareRobustCommand\onedot{\futurelet\@let@token\@onedot}
\newcommand{\@onedot}{\ifx\@let@token.\else.\null\fi\xspace}
\makeatother

\newcommand{\ie}{i.\,e.,\xspace}
\newcommand{\eg}{e.\,g.,\xspace}

\begin{document}
%
\title{Efficient Annotation of Medieval Charters}
%
%
\author{Anguelos Nicolaou\inst{1}\orcidID{0000-0003-3818-8718} \and
 Daniel Luger\inst{1}\orcidID{0000-0003-1778-4908} \and
Franziska Decker\inst{1}\orcidID{0000-0003-1778-4908} \and
Nicolas Renet\inst{2}\orcidID{0009-0007-5861-2578} \and 
Vincent Christlein\inst{2}\orcidID{0000-0003-0455-3799} \and
Georg Vogeler\inst{1}\orcidID{0000-0002-1726-1712}}
%
\authorrunning{Nicolaou et al.}
%
\institute{Center for Information Modeling (ZIM), University of Graz, \\Elisabethstraße 59/III, 8010 Graz, Austria\\
\email{anguelos.nicolaou@gmail.com} \email{firstname.lastname@uni-graz.at} \and
Pattern Recognition Lab, Friedrich-Alexander-Universität Erlangen-Nürnberg, 91058 Erlangen, Germany\\
\email{vincent.christlein@fau.de}
}
\maketitle              
\begin{abstract}
Diplomatics, the analysis of medieval charters, is a major field of research in which paleography is applied.
Annotating data, if performed by laymen, needs validation and correction by experts.
In this paper, we propose an effective and efficient annotation approach for charter segmentation, essentially reducing it to object detection.
This approach allows for a much more efficient use of the paleographer's time and produces results that can compete and even outperform pixel-level segmentation in some use cases.
Further experiments shed light on how to design a class ontology in order to make the best use of annotators' time and effort.
Exploiting the presence of calibration cards in the image, we further annotate the data with the physical length in pixels and train regression neural networks to predict it from image patches.

\end{abstract}

\input{introduction}
\input{layout}
\input{frat}
\input{resolution}
\input{conclusion}

\section*{Acknowledgements}
The work presented in this paper has been supported by ERC Advanced Grant project (101019327) "From Digital to Distant Diplomatics" and by the DFG grant No. CH 2080/2-1 "Font Group Recognition for Improved OCR".

\bibliographystyle{splncs04} 
\bibliography{bibliography}

\end{document}

%% file: introduction.tex
\section{Introduction}
\label{s:introduction}

\subsection{Diplomatics}
\label{ss:diplomatics}
Diplomatics is the study of diplomatic charters, medieval documents of a legal nature having a highly regular form.
While diplomatics is mostly focused on the content of the charters, many visual aspects are subject of investigation. 

\begin{figure}[t]
	\centering
	\includegraphics[width=.44\textwidth]{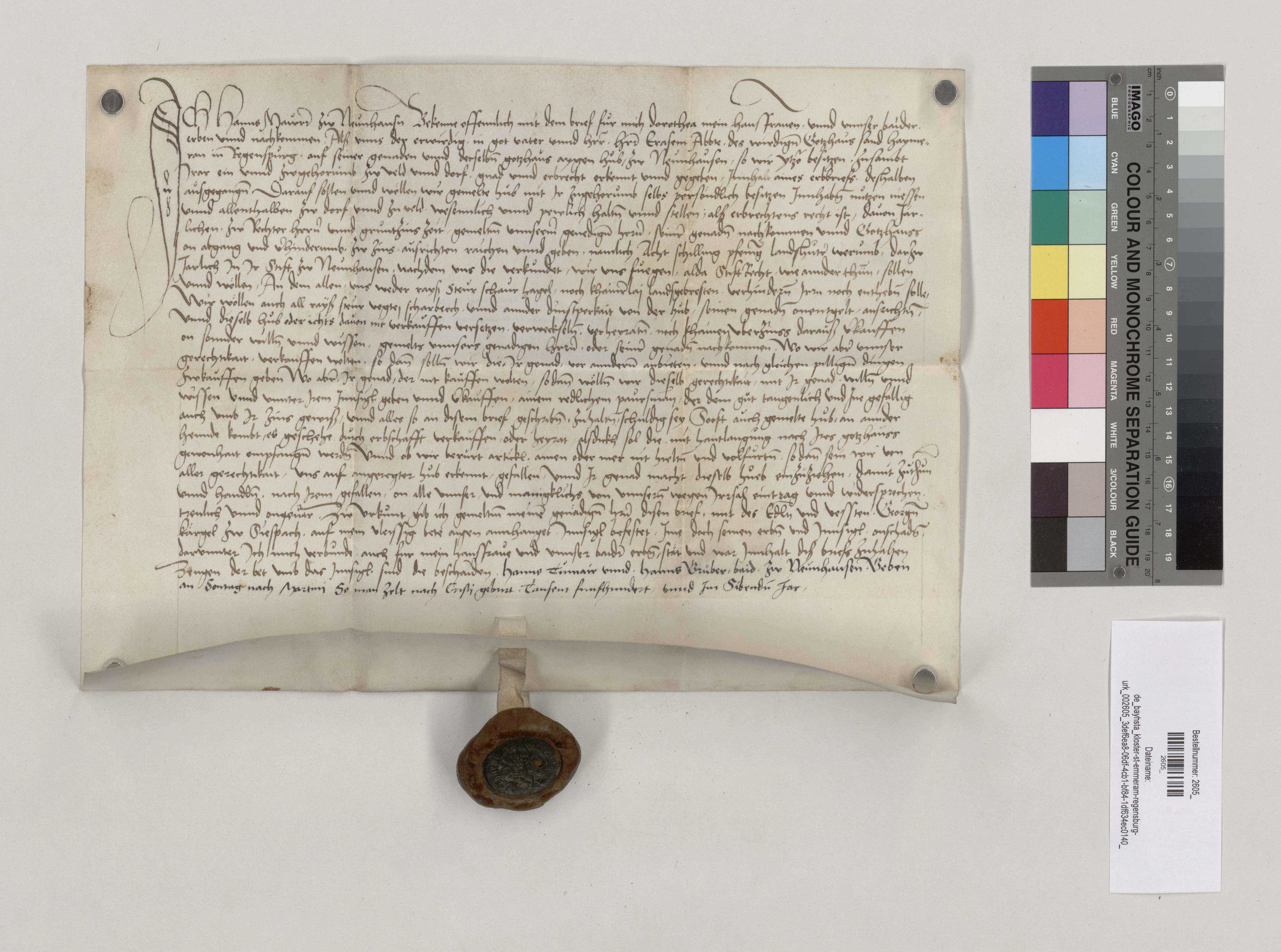} 
	\hspace{0.2cm}
	\includegraphics[width=.49\textwidth]{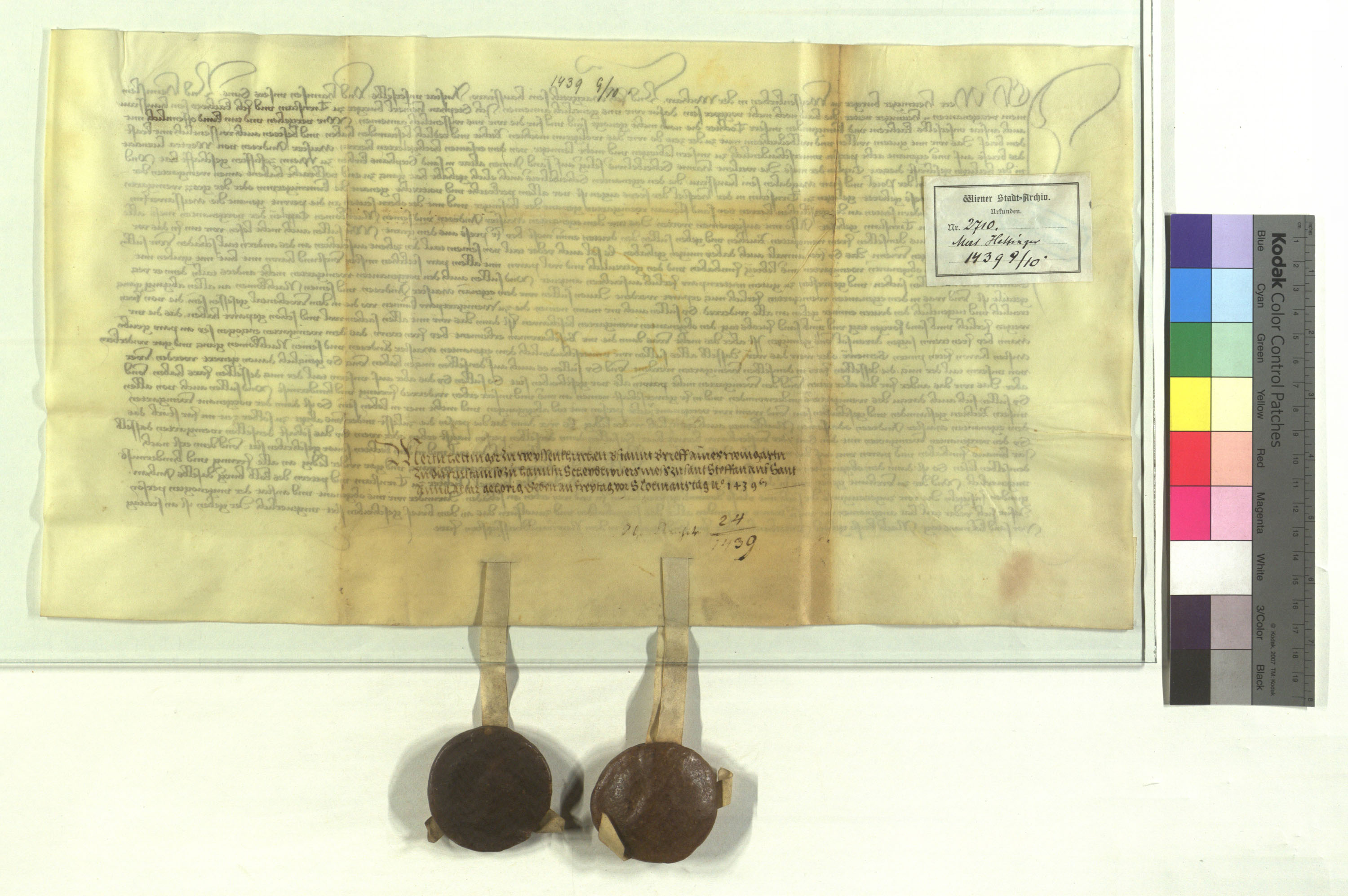}
	\caption{Left: a typical charter front (recto); right: a typical charter back-side on fine parchment (verso)}
	\label{fig:typical_charter}
\end{figure}

Sigillography (the study of diplomatic seals), forgery detection, document identification, dating, etc.\ are all  strongly dependent on analysing visual attributes of the documents making image analysis tools indispensable.
Although diplomatics data has been part of larger datasets of historical document images, pure image analysis on charters has not been that common.
Standing out among recent studies, Boro{\c{s}} et al.~\cite{borocs2020comparison} demonstrates
that Named Entity Recognition directly on images works better than with a two stage pipeline and the work by Leipert et al.~\cite{leipert2020notary} which addresses the problem of class imbalance in the context of image segmentation.
In \cref{fig:typical_charter} two typical charters, one verso and one recto, can be seen.

\subsection{Monasterium.net}
\label{ss:monasterium}
Our work is motivated by the need to analyse online archives, in particular \url{monasterium.net}, which constitutes the largest collection of continental European archives. It is focused on central Europe, containing 170 archives in total from Austria, Czech Republic, Estonia, Germany, Hungary, Italy, North Macedonia, Poland, Romania, Serbia, Slovenia, and Slovakia.
The data in such archives is highly inhomogeneous in nature and many details about the acquisition process are unavailable and can only be guessed at.
Thus, there is a need to perform automatic image analysis in order to obtain a proper overview of the quality of such archives. 

\subsection{Challenges}
\label{ss:chalenges}
At the same time, in traditional methods of diplomatics, 
much of the scientist's attention and effort is given to each specific charter.
In the late medieval period, charters grow to the hundreds of thousands, much more than a scholar can embrace.
This makes the need of automatic analysis assistance unavoidable.
Among humanists such automation is known by the term \textit{distant reading},
it is used to signify automatic text analysis instead of analysing the original documents; in contrast with the manual analysis of each particular document.
this notion also extends to visual analysis known as \textit{distant seeing}.
Distant reading and seeing, as opposed to typical image analysis, is practically impossible to scrutinise experimentally and the extent to which such methodologies are falsifiable is a nuanced discussion.
These epistemological challenges mandate a very good understanding of the underlying data and biases in the employed methods.
From an engineering perspective, we considered that the most precious resource
is the diplomatist's time and most decisions were made by trying to minimize the scholar's time needed for annotation and then trying to maximize the use of the spent time.

The main contributions of this paper are as follows:
\begin{enumerate}
	\item We introduce a highly efficient data annotation tool called FRAT.\footnote{Software available under an open-source licence at \url{https://github.com/anguelos/frat} and in pypi \url{https://pypi.org/project/frat/}}
	\item We present an object detection diplomatics dataset and report experiments using it.
 \footnote{The code has been forked from the \href{https://github.com/ultralytics/yolov5}{original YOLOv5} repository, the specific scripts used for the experiments are in the bin folder \url{https://github.com/anguelos/yolov5/tree/master/bin}}
	\item We present a dataset with manually annotated image resolution estimates and present a model that learns to infer them.\footnote{The code used to run the resolution regression experiments, including links to the datasets, is available in a github repository \url{https://github.com/anguelos/resolution_regressor}} 
\end{enumerate}

%% file: layout.tex
\section{Layout Analysis}
\label{s:layout}

\subsection{1000 Charter Dataset}
\label{ss:1000charters}
\begin{figure}[t]
	\centering
	\includegraphics[width=\textwidth]{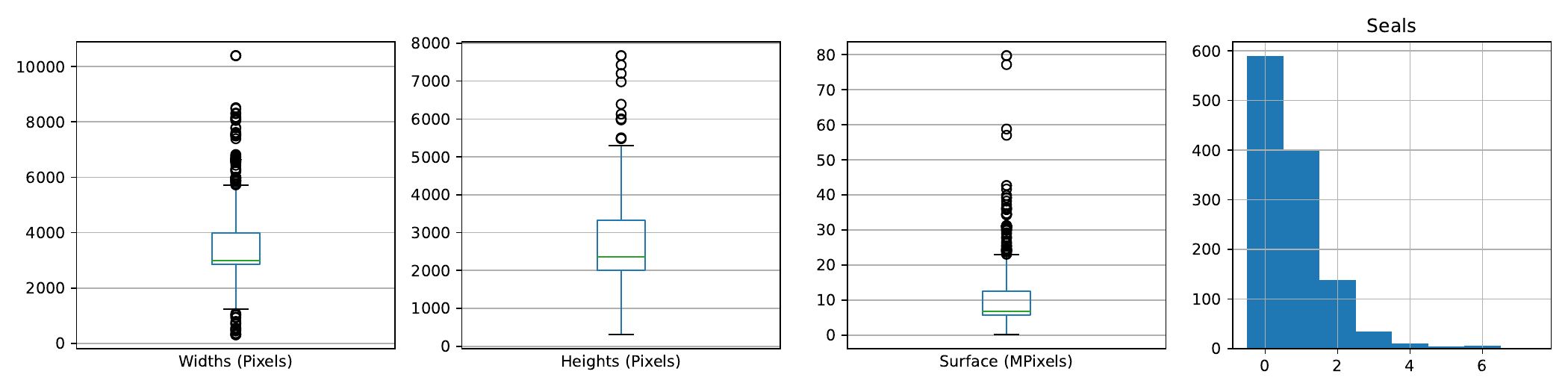}
	\caption{Distributions of image cardinalities in the 1000-charters dataset, and a histogram of occurring seals per image}
	\label{fig:1000charters}
\end{figure}

In order to make sure we gain a proper understanding of the data available in \url{monasterium.net}, we randomly chose 1000 charters from it.
The heterogeneous nature of the dataset is illustrated by \cref{fig:1000charters}, in which we plot the distribution of width, height, and surface in pixels and mega-pixels, respectively.
Additionally, the number of diplomatic seals occurring in each charter image is given.

The dataset not only aims at understanding the quality and nature of the diplomatics data in monasterium.net but also at supporting the creation of pipelines that will offer distant viewing and distant reading.
Specifically, we model the layout analysis problem as an object detection problem.
As our goal is to maximize the efficiency of diplomatic scholars, we devised an annotation strategy that would require as little time as possible per charter.
Diplomatic charters are considered highly formulaic~\cite{filatkina2018historische}, commonly containing a single dominant text-block as can be seen in \cref{fig:typical_charter}.
Since the images do not seem to need any dewarping, we modeled all layout annotation tasks as defining two-point rectangles, \ie a rectangle extending from the top left corner to the bottom right corner. 
Each rectangle belongs to one of 11 specific classes.
While these classes are from the point of view of annotation flat and independent, they imply a hierarchical structure:
\begin{itemize}
\item \textbf{\textit{"No Class"}}: Reserved, should never occur normally.
\item \textbf{\textit{"Ignore"}}: Used for possible detections that should be tolerated but not required when doing performance evaluation. 
Might make sense to exclude this class of objects when training a model, \eg weights placed on charters in order to hold it in place for photography. 
\item \textbf{\textit{"Img:CalibrationCard"}}: Calibration cards are usually the 21 cm Kodak ones but not always; they usually occur on both sides of the charter.
\item \textbf{\textit{"Img:Seal"}}: Many charters contain one or more seals. Seals are of particular interest for sigillographers, i.e. a small sub-community of paleographers focused on the study of seals. 
\item \textbf{\textit{"Img:WritableArea"}}: This class represents the background material on which a charter was written. It consists typically of parchment but paper is also common in late medieval times. The following classes can only reasonably occur inside this class.
\item \textbf{\textit{"Wr:OldText"}}: This class represents the typically single text block of the charter. This rectangle should aim to contain the whole tenor, if possible only the tenor of the charter. 
\item \textbf{\textit{"Wr:OldNote"}}: This refers to pieces of text put there during the charter's creation/early life that are not part of the tenor. 
\item \textbf{\textit{"Wr:NewText"}}: This refers to pieces of text added at later times, typically by archivists. From a legal point of view it should be considered irrelevant.
\item \textbf{\textit{"Wr:NewOther"}}: This refers to items added to the charter after its creation that are not purely text, \eg stamps and seal imprints.
\item \textbf{\textit{"WrO:Ornament"}}: This refers to items put on the charter at creation time for principally aesthetic purpose.
\item \textbf{\textit{"WrO:Fold"}}: This class is meant to indicate a typical charter element, the plica. 
A fold that is not strictly a plica but is of significant size is still considered a fold.
\end{itemize}
As the class names imply: calibration cards, seals, and writable areas can be perceived as children of the root element ``image'', and all other classes are children of the writable area.
This structural interpretation is suggested but not enforced in any way.

\subsection{Object Detection}
\label{ss:yolo}
As an object detector we employed an off-the shelf YOLOv5~\cite{Jocher_YOLOv5_by_Ultralytics_2020},
which was only modified to include fractal-based augmentations, that are especially well suited for document images~\cite{nicolaou2022tormentor}
.
The small model variant was chosen, which consists of 7.2M parameters at half precision and it was trained on images resized to \numproduct{1024x1024} pixels.
The dataset contains 1184 annotated images, which were split into 741 images for the train set and 440 images for the validation set.

\begin{figure}
    \centering
    \includegraphics[width=.8\textwidth]{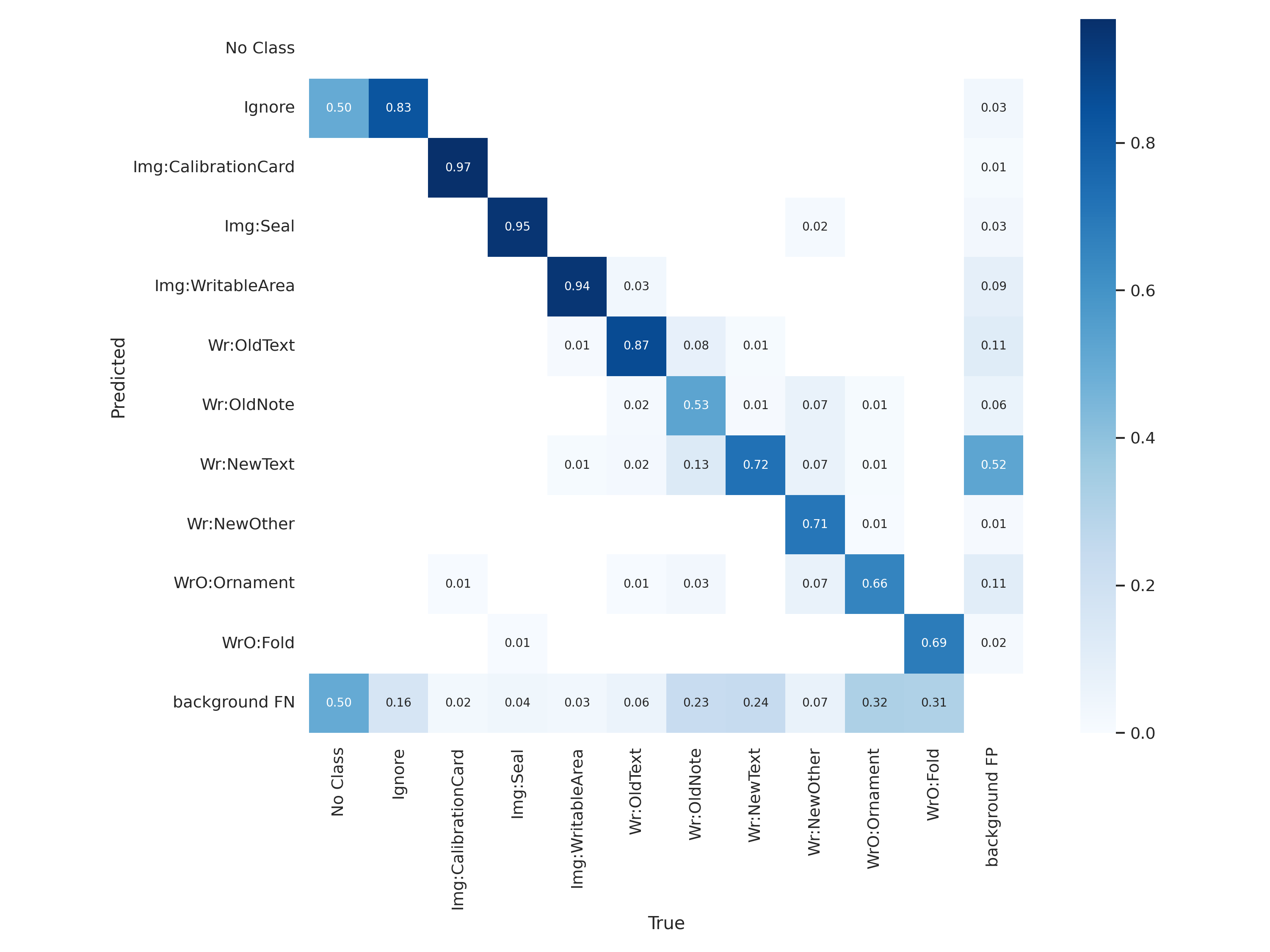}
		\caption{Confusion Matrix for all classes at \SI{50}{\percent} IoU}
    \label{fig:yolo_confusion}
\end{figure}

The mean average precision (mAP) for all classes at \SI{50}{\percent} Intersection over Union (IoU) obtained is \SI{73.21}{\percent}, while at \SI{95}{\percent} IoU it is \SI{52.72}{\percent}.
These numbers demonstrate very good results, for some classes, the results are even nearly flawless. 
The confusion matrix is shown in \cref{fig:yolo_confusion}.
We can see that classes \textbf{\textit{Img:Seal}}, \textbf{\textit{Img:WriteableArea}}, and \textbf{\textit{Img:CalibrationCard}} all score at \SI{94}{\percent} precision or higher.
These classes are the ones for which annotators only need little interpretation.
Class \textbf{\textit{"Wr:OldText"}} is also performing fairly well with an accuracy of \SI{87}{\percent}, especially if one takes into account that \SI{8}{\percent} of missed objects are detected as \textbf{\textit{"Wr:OldNote"}}, which indeed can easily be confused.
This class is also quite important as it is the one that detects text blocks, 
which can subsequently be segmented into text lines so that they can be passed to a handwritten text recognition (HTR) process.

\begin{figure}
    \centering
    \includegraphics[width=\textwidth]{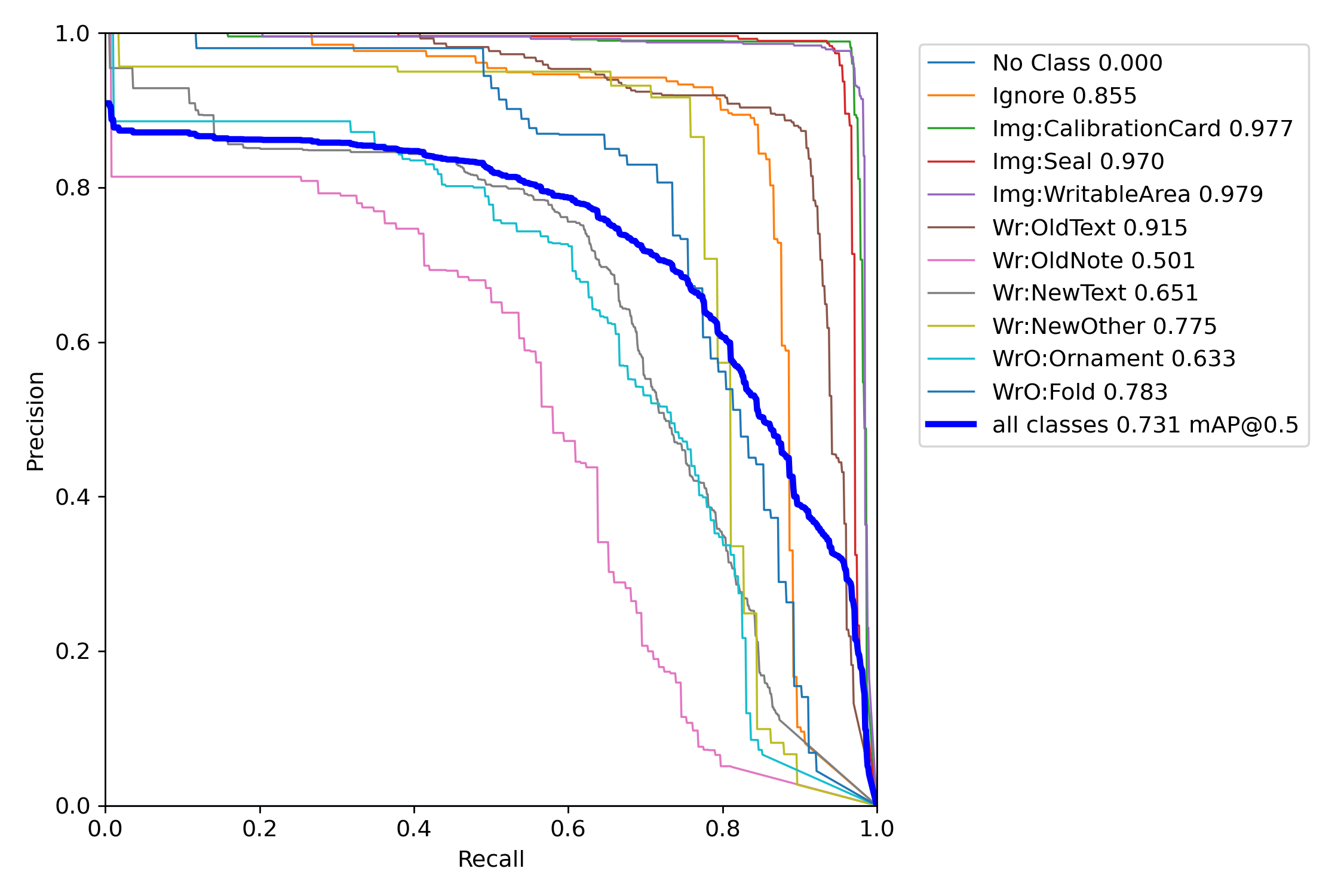}
		\caption{Precision Recall curves for all classes at \SI{50}{\percent} IoU}
    \label{fig:yolo_pr}
\end{figure}

\cref{fig:yolo_pr} shows the precision recall curves of each class along with their mAP at \SI{50}{\percent} IoU.
It can be observed that the mAP scores are significantly higher than the precision values reported in the confusion matrix.

%% file: frat.tex
\subsection{Annotations with FRAT}
\begin{figure}[t]
\centering
\includegraphics[width=.45\textwidth, height=.35\textwidth]{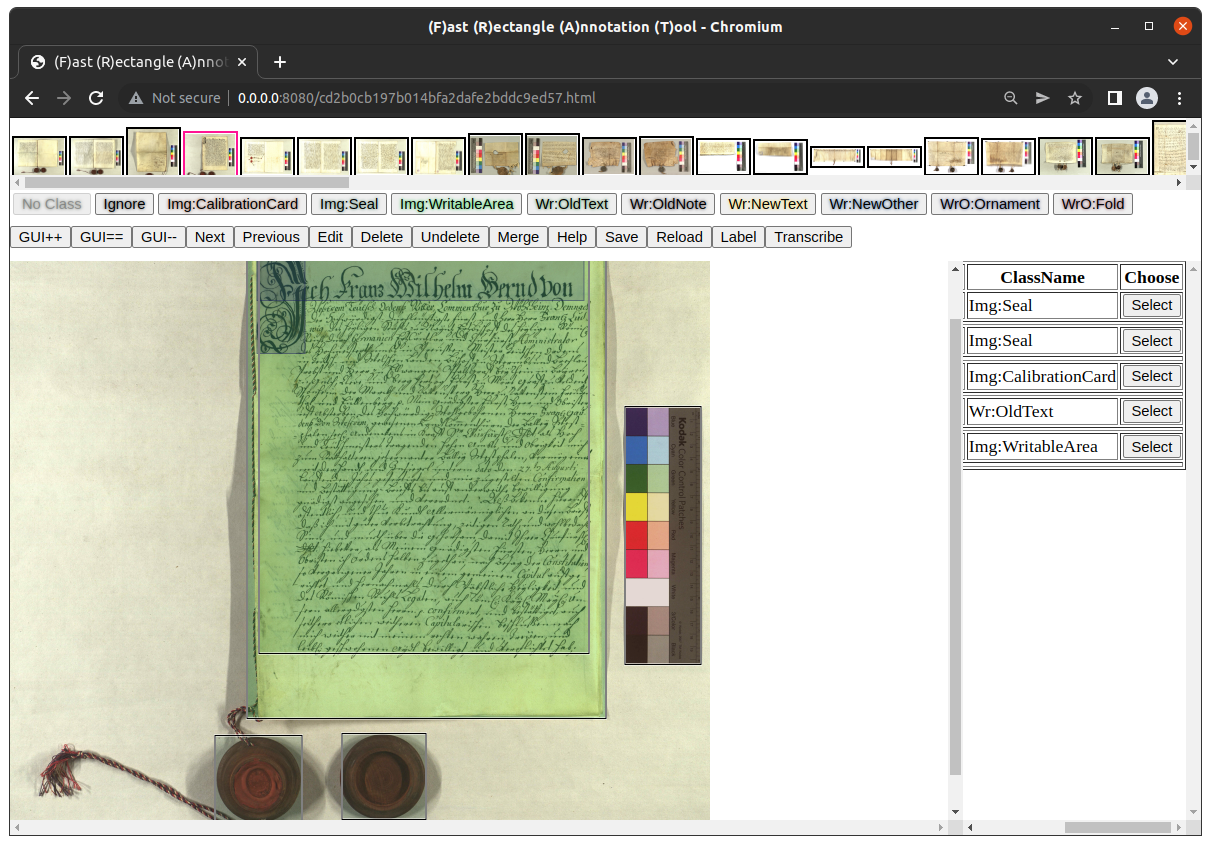} 
\quad\includegraphics[width=.45\textwidth,  height=.35\textwidth]{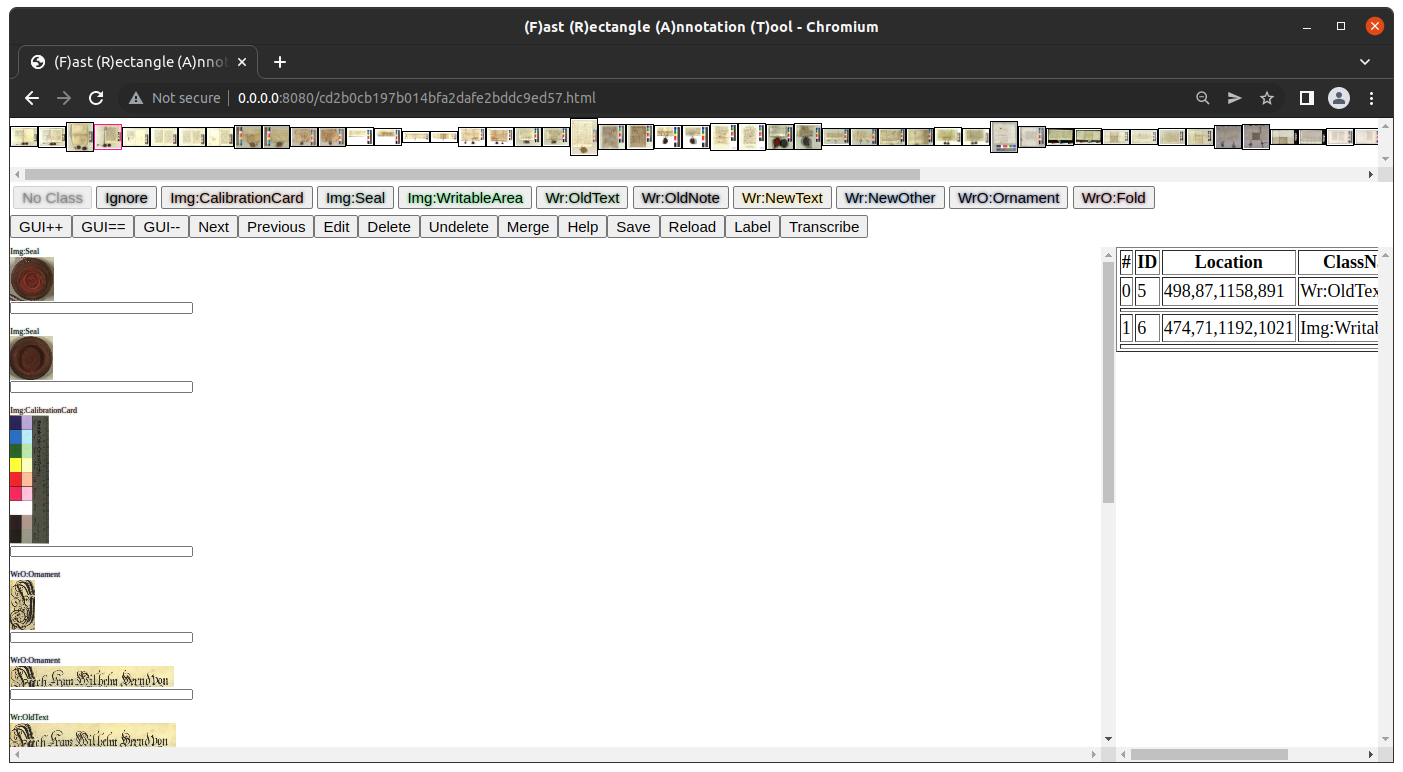}
\caption{FRAT's ``Label'' (left) and ``Transcription'' (right) modes.}
\label{fig:frat}
\end{figure}

We introduce a lightweight annotation tool for document image annotation.
The tool can annotate rectangles by just two-points with an arbitrary set of classes.
Other than a class, each rectangle can also have an associated Unicode transcription as well as a comment.
The tool is called Fast Rectangle Annotation Tool (FRAT).
It is web-based and written in Python and Javascript and therefore runs on Linux and Mac OsX natively, and on Windows using Docker.
FRAT is available in pypi and can be easily deployed with pip.
In order to accommodate this very generic data model, the tool has its own JSON-based file format reflecting these fundamental assumptions and nothing more.
The User Interface (UI) is aimed at being extremely lightweight and fast, to a certain extent at the expense of precision.
It is inspired by gaming interfaces, where all buttons are
accessible as single keys on the keyboard, see \cref{fig:frat}. 
Set operations such as rectangle union and subtraction allow the annotator to figure out efficient use patterns that best suit the domain of data employed.

A user-overridable JSON configuration file defines all user preferences as well as the class labels to be employed.
All that is needed in order to begin a new project is to define the class labels in the config file.
Class labels are considered by FRAT to be flat although they might be perceived to form a hierarchy 
For example, if a class is \emph{textline} and another class is\emph{word}, FRAT will ignore the fact that a textline consists of words but simple heuristics could be made to realise this hierarchical relation,
but this hierarchical relation could be realized through fairly straightforwrd post-processing heuristics, that would create textlines out of word bounding boxes.
Custom exporting to PAGE-XML~\cite{pletschacher2010page} for text-styled text-lines is demonstrated in the code. 
Since object detection fits best, the scenario FRAT was built around, generic exporting to the MS-COCO dataset format~\cite{lin2014microsoft}. 
YOLOv5~\cite{yolov5software} is also available but no transcriptions are exported.

From a user interface perspective, FRAT features two modes of image annotation:
The first mode, ``Label'', works like a typical generic object annotation tool, with an extra option of transcribing or commenting each annotated object.
Defining an object is done by a single drag along its diagonal.
Single-key shortcuts allow for editing, manipulating, and navigating through all defined objects.
The second mode, ``Transcribe'', is aimed at transcribing text and reviewing object classes.
In this mode a list of all defined objects, their classes, their transcriptions, and their comments is displayed in a scrollable list.
This mode is most effective when annotating transcriptions of textlines or words, since allowing to transcribe something out of its context might make the groundtruthing more realistic not allowing annotators to use high-level semantic cues from the nearby text.
FRAT has also been used to annotate data including class and transcriptions in~\cite{seuret2023combining}.

%% file: resolution.tex
\section{Resolution Prediction}
\label{s:resolution}
Given that many images of diplomatic charters contain calibration cards, it is possible to infer an image resolution approximation.
This is quite useful as it allows for quality analysis of large archives but would also allow to associate visual objects of the charter seal, plica, etc.\ with physical world sizes.
As is often the case, the data we are dealing with has been acquired in a non-standardized way and any information about the acquisition process is considered irrecoverable.
Under the assumption that during acquisition the charter, the calibration card, and the seals lie on the same plane and that local depth deviations are negligible, we can use the marked distances on the calibration cards to get an estimate for the width and height of any pixel that lies on that plane.
Assuming a view angle $\phi$, and assuming that calibration cards are predominantly on the horizontal edge of the cameras field of view, the estimated pixel resolution lies in between $1$ and $\cos(\frac{\phi}{2})$ of the estimated pixel resolution.
In the case of a \SI{50}{\milli\meter} focal length, this error would be at most \SI{7.62}{\percent} while in the case of \SI{35}{\milli\meter} focal length, it would be at most \SI{13.4}{\percent}.
As the images have been cropped, and nothing is known about this process, this distortion is not reversible but can be used to provide some bounds.


In order to create the ground truth data for the task of resolution estimation, we used the object detector to isolate the calibration cards, writeable areas, and seals.
For simplicity's sake, we used FRAT and marked rectangles with a \SI{5}{\centi\meter} diagonal wherever that was possible.
When \SI{5}{\centi\meter} was not an option due to image size, we marked \SI{1}{\centi\meter}.
In the few cases where there was no visible inch or cm marking, we operated on the assumption that inches are divided by halves, quarters, and eights; while centimeters are divided by halves and millimeters.
For all analyses, we used Pixel per Centimeter (PpCm) unit.

\subsection{Baseline}
For a baseline, we considered the calibration-card object detector described in \cref{s:layout}.
Specifically, since almost all calibration cards are variants of the Kodak calibration cards, they have a length of \SI{20.5}{\centi\meter}, even if they differ in other details, while having a height of less than \SI{5}{\centi\meter}.
The calibration cards occur always on the sides of the images vertically or horizontally aligned either vertically or horizontally, by choosing the largest side of the bounding box, we can with high certainty obtain an estimate on how many pixels make a length of 20.5 cm on the photographed plane.
This heuristic interpretation of the calibration card object detection gave a Spearman's correlation of 89.13\% and makes it a reasonably strong baseline.
To the knowledge of the authors there is no other baseline available in the literature that is applicable in this experimental context.

\subsection{ResResNet}
We introduce a network called ResResNet
which is a typical ResNet~\cite{he2016deep} where the final fully connected layer is substituted with a small multi-layer perceptron (MLP) consisting of three fully connected layers with 512, 256, and 1 nodes. 
While the ResResNet can utilise any ResNet architecture as a backend, we mostly experimented with ResNet18 in order to limit the capacity of the model to suit our relatively small dataset.
Due to their average pooling layer, ResResNet can operate on images of arbitrary size.
Image resolution can be considered as a texture-base analysis. 
Thus, during inference, we run ResResNet on a sliding window over the initial image and consider the median prediction as a result for the image. Note: in \cref{fig:correlations} this inference method is called stable.

\begin{figure}[t]
    \centering
    \includegraphics[width=.8\textwidth]{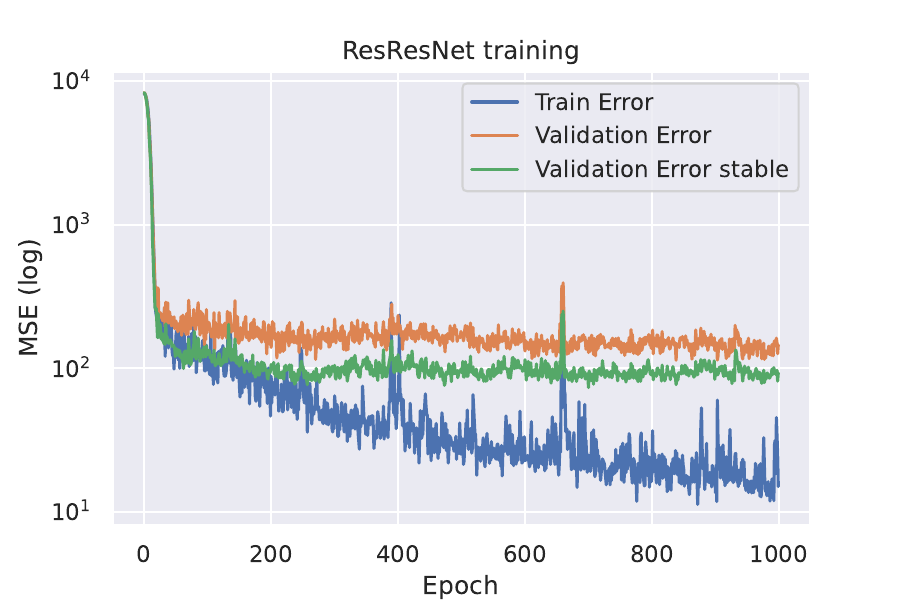}
    \caption{ResResNet training convergence}
    \label{fig:resresnet_training}
\end{figure}

\subsection{Training}
We used cropped images of the writeable areas as detected with the object detector for inputs and the estimated PpCm as deduced from the annotations on the respective calibration cards.
For training, each input image was randomly cropped to a square of \numproduct{512x512} pixels providing sufficient data augmentation.
We employed a Mean Square Error (MSE) loss function, a learning rate of 0.0001 and used an ADAM optimiser.
The batch-size was set to 16.
The backbone of the ResResNet was initialised with weights pre-trained on Imagenet~\cite{deng2009imagenet} and would not converge at all when initializing with random weights.
In total, we trained the model for 1000 epochs.
\Cref{fig:resresnet_training} shows that the model converges to its validation performance after epoch 200 while the training error continues to decrease, it is evident that no problematic overfitting occurs at least for the 1000 epochs.

\subsection{Results and Analysis}
\begin{figure}[t]
\centering
\subcaptionbox{\label{corr_a}}{
\includegraphics[width=.3\textwidth]{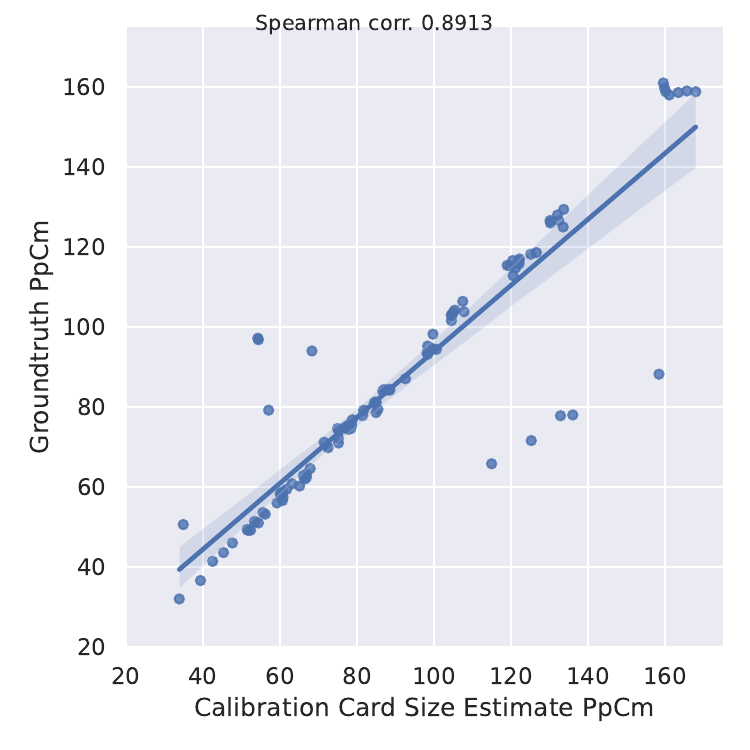}
}
\subcaptionbox{\label{corr_b}}{
\includegraphics[width=.3\textwidth]{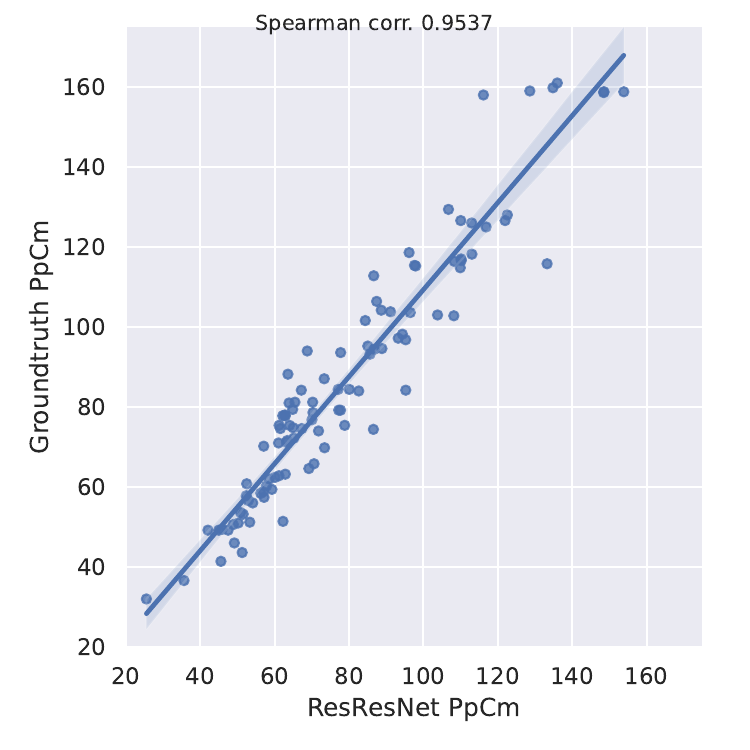} 
}
\subcaptionbox{\label{corr_c}}{
\includegraphics[width=.3\textwidth]{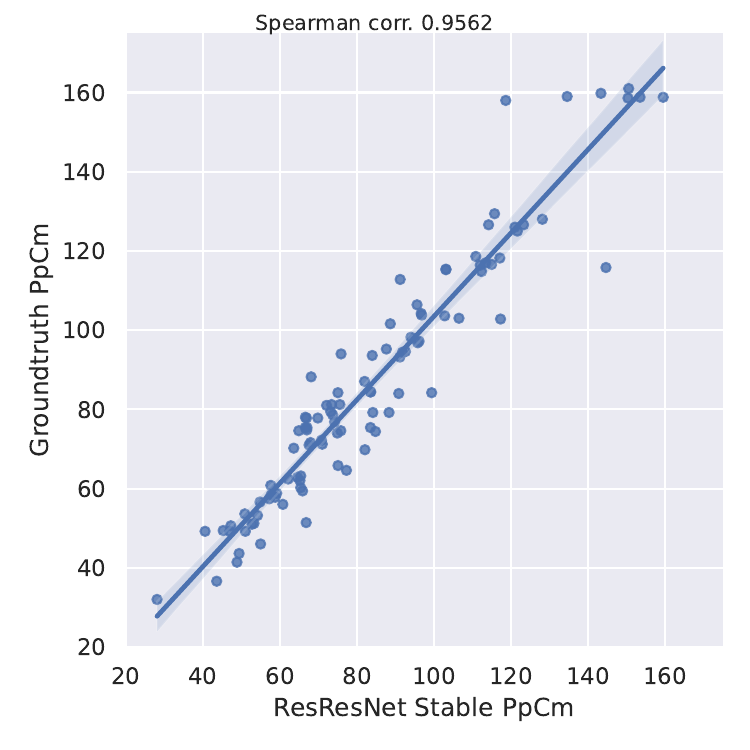}
}
\caption{Scatter plots of the resolution estimates with the ground-truth for the validation set. The resolution estimate from object detection on \subref{corr_a} calibrations cards, \subref{corr_b} the resolution estimate from ResResNet on the cropped charter, and \subref{corr_c} the resolution estimate from ResResNet on the cropped charter after outlier removal.}
\label{fig:correlations}
\end{figure}

Resolution estimation is a regression task. 
Hence, typical performance metrics such as Accuracy and F-score are not applicable.
For our performance analysis, we use MSE and and we consider Spearmann's correlation to be more appropriate as the errors do not follow a Gaussian distribution in a very loose sense. 
MSE quantifies the size of mistakes in total while Spearman's correlation quantifies how often predictions are correct.
\Cref{fig:correlations} shows the results.
While both validation and train sets were curated such that they would only refer to images that contained a calibration card out of necessity, they fail to demonstrate the base limitation of the baseline system, which is that a calibration card is required for an estimate to be possible at all.

\begin{figure}[t]
\centering
\subcaptionbox{\label{inliers_a}}{
	\includegraphics[width=.45\textwidth]{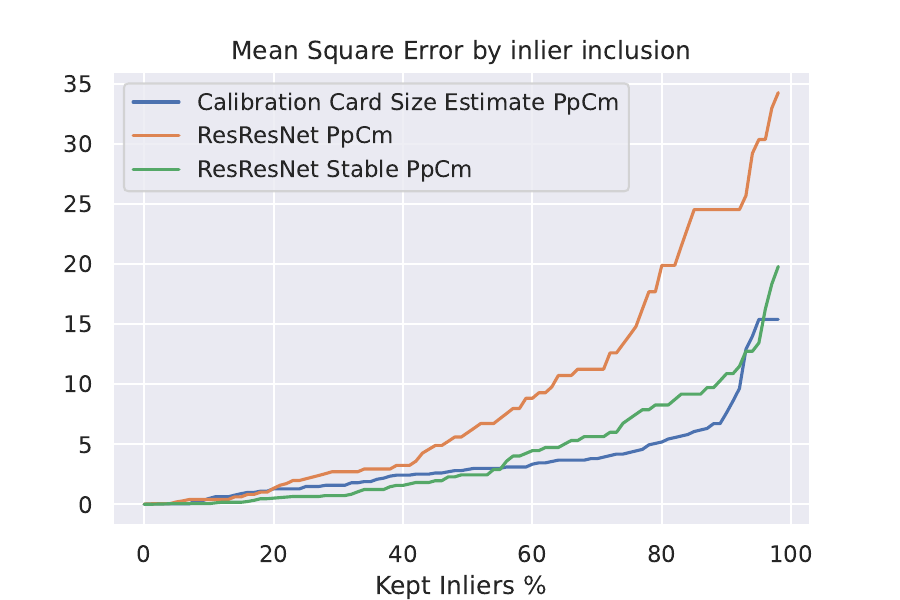}
}
\subcaptionbox{\label{inliers_b}}{
\includegraphics[width=.45\textwidth]{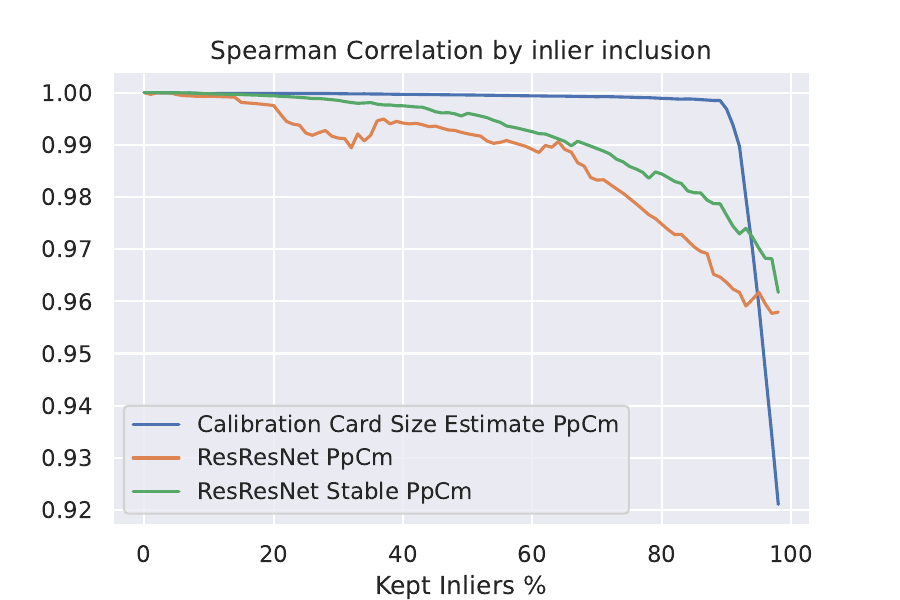}
}
\caption{Outlier contribution to PpCm prediction. \subref{inliers_a} MSE growth and \subref{inliers_b} Spearmann's correlation growth as more outliers are included.}
\label{fig:inliers}
\end{figure}

Therefore, we gradually remove samples from the validation set from the most erroneous one (according to the squared error) to the least erroneous one.
The results are shown in \cref{fig:inliers}.
We notice that the calibration card baseline performs the best for most samples but overall ResResNet is more reliable as it handles hard samples more gracefully. 
Furthermore, ResResNet deals with images that do not contain any calibration card at all.
The fact that ``texture'' analysis is enough to learn to predict physical resolution indicates that historical textual data is by nature scale-sensitive and because the trained image analysis models operate on various scales, we can speculate that they implicitly learn multiple scale representations of the same phenomenon.
Further experiments that might help validate these hypotheses would be to see whether re-scaling train and test sets for typical textural image analysis tasks such as binarization would improve performance on these tasks.
Another approach to better make use of this hypothesis would be to add ResResNet as a Spatial Transformer Network (STN)~\cite{NIPS2015_33ceb07b} with PpCm being the single degree of freedom.


%% file: conclusion.tex
\section{Discussion and Conclusion}
\subsection{Data Annotation Efficiency}
In this paper, we present a work on Diplomatic charters motivated by making optimal use and re-use of domain experts.
Our principal research question was how much effort can we economise on annotating and how far will these weak annotations take us in the use of subsequent pipelines.
While observing the annotation task, it was evident that marking the seals on a charter would only take a few seconds while waiting for the image to load might take quite a bit longer.
To a great extent the proposed class ontology was designed to make better use of the scholar's time by trying opportunistically to ask more questions while they wait.
Essentially we tried to increase the objects to be annotated on each image in order to dilute waiting time of annotators to the extent where it would not cause frustration.

The defined classes were chosen not only on the theoretical completeness but also on what is easy to obtain and what might be easier to learn from a machine learning perspective.
Modeling the classes and the annotation recommendation benefited greatly from the interdisciplinary nature of the team.
It is worth pointing out that the class that made diplomatists sceptical was the class \textbf{\textit{Wr:Fold}}. Since it does not coincide totally with the established ``plica'' charter element. 
It is also the worst performing one, while this could also be interpreted with the fact that folds are not well defined in rectangles and therefore object detection might possibly not be well suited for these elements.

\subsection{Towards a Full Diplomatics Pipeline}
It also appears that in the case of diplomatics, there is no need for a dewarping stage as in other DIA (Document Image Analysis) pipelines~\cite{neudecker2019ocr}. 
While dewarping is a very simple step on its own, it complicates a DIA pipeline quite a bit because pixel coordinates are not consistent across the whole pipeline and from a user interface perspective it becomes a major challenge to align localised outputs to the original images.
The performance trade-off of using simple object detection as the only supervision data for layout analysis in other use cases has yet to be determined. 
Thus, there is strong evidence that it is sufficient in the diplomatics case.

\subsection{Future Work}
The work presented in this paper is mostly about presenting a preliminary visual analysis of a large corpus of charters. 
Several questions for further research are still open.
The most interesting finding from a DIA perspective is the fact that, in the case of diplomatic charters, we can predict the PpCm of a charter by looking at patches of \numproduct{512x512} pixels.
An open question is whether resolution can be predicted because of the features of the foreground, the background (parchment), or the acquisition process and how we can exploit that.
The other direction in which the annotated data can be used is in high quality synthesis of data for training the subsequent stages of the pipeline such as binarization and HTR.

%% file: paper.bbl
\begin{thebibliography}{10}
\providecommand{\url}[1]{\texttt{#1}}
\providecommand{\urlprefix}{URL }
\providecommand{\doi}[1]{https://doi.org/#1}

\bibitem{borocs2020comparison}
Boro{\c{s}}, E., Romero, V., Maarand, M., Zenklov{\'a}, K.,
  K{\v{r}}e{\v{c}}kov{\'a}, J., Vidal, E., Stutzmann, D., Kermorvant, C.: A
  comparison of sequential and combined approaches for named entity recognition
  in a corpus of handwritten medieval charters. In: 2020 17th International
  conference on frontiers in handwriting recognition (ICFHR). pp. 79--84. IEEE
  (2020)

\bibitem{deng2009imagenet}
Deng, J., Dong, W., Socher, R., Li, L.J., Li, K., Fei-Fei, L.: Imagenet: A
  large-scale hierarchical image database. In: 2009 IEEE conference on computer
  vision and pattern recognition. pp. 248--255. Ieee (2009)

\bibitem{filatkina2018historische}
Filatkina, N.: Historische formelhafte sprache. In: Historische formelhafte
  Sprache. de Gruyter (2018)

\bibitem{he2016deep}
He, K., Zhang, X., Ren, S., Sun, J.: Deep residual learning for image
  recognition. In: Proceedings of the IEEE conference on computer vision and
  pattern recognition. pp. 770--778 (2016)

\bibitem{NIPS2015_33ceb07b}
Jaderberg, M., Simonyan, K., Zisserman, A., kavukcuoglu, k.: Spatial
  transformer networks. In: Cortes, C., Lawrence, N., Lee, D., Sugiyama, M.,
  Garnett, R. (eds.) Advances in Neural Information Processing Systems.
  vol.~28. Curran Associates, Inc. (2015)

\bibitem{Jocher_YOLOv5_by_Ultralytics_2020}
Jocher, G.: {YOLOv5 by Ultralytics} (May 2020). \doi{10.5281/zenodo.3908559},
  \url{https://github.com/ultralytics/yolov5}

\bibitem{yolov5software}
Jocher, G., Stoken, A., Borovec, J., NanoCode012, ChristopherSTAN, Changyu, L.,
  Laughing, tkianai, Hogan, A., lorenzomammana, yxNONG, AlexWang1900, Diaconu,
  L., Marc, wanghaoyang0106, ml5ah, Doug, Ingham, F., Frederik, Guilhen,
  Hatovix, Poznanski, J., Fang, J., Yu, L., changyu98, Wang, M., Gupta, N.,
  Akhtar, O., PetrDvoracek, Rai, P.: {ultralytics/yolov5: v3.1 - Bug Fixes and
  Performance Improvements} (Oct 2020). \doi{10.5281/zenodo.4154370}

\bibitem{leipert2020notary}
Leipert, M., Vogeler, G., Seuret, M., Maier, A., Christlein, V.: The notary in
  the haystack--countering class imbalance in document processing with cnns.
  In: Document Analysis Systems: 14th IAPR International Workshop, DAS 2020,
  Wuhan, China, July 26--29, 2020, Proceedings 14. pp. 246--261. Springer
  (2020)

\bibitem{lin2014microsoft}
Lin, T.Y., Maire, M., Belongie, S., Hays, J., Perona, P., Ramanan, D.,
  Doll{\'a}r, P., Zitnick, C.L.: Microsoft coco: Common objects in context. In:
  European conference on computer vision. pp. 740--755. Springer (2014)

\bibitem{neudecker2019ocr}
Neudecker, C., Baierer, K., Federbusch, M., Boenig, M., W{\"u}rzner, K.M.,
  Hartmann, V., Herrmann, E.: Ocr-d: An end-to-end open source ocr framework
  for historical printed documents. In: Proceedings of the 3rd international
  conference on digital access to textual cultural heritage. pp. 53--58 (2019)

\bibitem{nicolaou2022tormentor}
Nicolaou, A., Christlein, V., Riba, E., Shi, J., Vogeler, G., Seuret, M.:
  Tormentor: Deterministic dynamic-path, data augmentations with fractals. In:
  Proceedings of the IEEE/CVF Conference on Computer Vision and Pattern
  Recognition. pp. 2707--2711 (2022)

\bibitem{pletschacher2010page}
Pletschacher, S., Antonacopoulos, A.: The page (page analysis and ground-truth
  elements) format framework. In: 2010 20th International Conference on Pattern
  Recognition. pp. 257--260. IEEE (2010)

\bibitem{seuret2023combining}
Seuret, M., van~der Loop, J., Weichselbaumer, N., Mayr, M., Molnar, J., Hass,
  T., Kordon, F., Nicolau, A., Christlein, V.: Combining ocr models for reading
  early modern printed books (2023)

\end{thebibliography}
